\documentclass{article}
\usepackage{spconf,amsmath,graphicx}
\usepackage{hyperref}
\usepackage{xcolor}
\usepackage{multirow}

\usepackage{enumitem}
\def\model{BESTOW }

\title{BESTOW: Efficient and Streamable Speech Language Model with the Best of Two Worlds in GPT and T5 }

\name{
\begin{tabular}{c} 
Zhehuai Chen, He Huang,  Oleksii Hrinchuk, Krishna C. Puvvada, Nithin Rao Koluguri, Piotr Żelasko, \\
Jagadeesh Balam, Boris Ginsburg
\end{tabular}
}

\address{NVIDIA, Santa Clara, CA, USA}
\begin{document}

\maketitle

\begin{abstract}    
    %TODO(zhehuai): rewrite to show efficient, streamable, strong multitask results as the three focuses. 
    Incorporating speech understanding capabilities into pretrained large-language models has become a vital research direction (SpeechLLM). 
    The previous architectures can be categorized as:   
    i) {\em GPT-style}, prepend speech prompts to the text prompts as a sequence of LLM inputs like a decoder-only model; 
    ii) {\em T5-style}, introduce speech cross-attention to each layer of the pretrained LLMs. 
        We propose BESTOW architecture
    to bring {\em the BESt features from TwO Worlds} into a single model that is highly efficient and has strong multitask capabilities.
     Moreover, there is no clear streaming solution for either style, especially considering the solution should generalize to speech multitask.
 We reformulate streamable SpeechLLM as a read-write policy problem and unifies the offline and streaming research with BESTOW architecture.  
    Hence we demonstrate the first open-source SpeechLLM solution that enables {\em Streaming} and  {\em Multitask at scale} (beyond ASR) at the same time. 
    This streamable solution achieves very strong performance on a wide range of speech tasks (ASR, AST, SQA, unseen DynamicSuperb). It is end-to-end optimizable, with {\em lower training/inference cost}, and demonstrates LLM knowledge transferability to speech. 
\end{abstract}

\section{Introduction}

With the huge success of large language models (LLMs)~\cite{brown2020gpt,team2023gemini}, researchers start to explore the possibilities of extending the capabilities of LLMs with multi-modal understanding skills, and many works have been proposed to support image and audio understanding~\cite{alayrac2022flamingo,liu2024llava,zhang2023speechgpt,gong2023ltu-as,tang2023salmonn}.

This work focuses on leveraging speech encoder and LLM ({\em SpeechLLM}) to build a speech foundational model  for  many speech-and-audio to text applications (STT). 
One popular framework in the direction is Speech-LLaMA~\cite{wu2023decoder,fathullah2023prompting} and its extensions, which 
updates the input of LLM by prepending speech prompts to the text prompts while keeping the rest of LLM unchanged or LoRA finetuned. This design shows good modularity which allows knowledge transfer from LLMs to speech and results in strong ASR and AST performance~\cite{wang2023slm,qwen2023}. 
The modular design also shows strong in-context learning ability 
\cite{chen2024salm}. 
\cite{tang2023salmonn} further introduces a bespoke Q-former module before prepending speech prompts to better bring speech, audio, and music features to LLM space.

Nevertheless, there are several potential drawbacks in this popular design:
i) Efficiency problem raised from the interaction between self-attention and the longer speech embeddings than text, which will be elaborated in Section~\ref{sec:efficiency}. Workarounds like massive downsampling of speech embeddings usually come with information loss and cannot completely avoid. % the per-layer and per-step LLM computation increase from speech. 
%ii) Scalability. Partly because of above, there is limited success in scaling this type of design to very large scale, except in~\cite{wang2023slm,qwen2023}.
ii) As the speech embeddings of the whole utterance are treated as prompt and always prepended beforehand, it disables many streaming applications in speech, e.g. streaming ASR and simultaneous speech translation (SST). 
% i) efficiency; will analyze, resultant massive downsampling and information loss
% ii) scalable, essential for speech foundational model; train on large scale data
% iii) if the speech embedding is treated as prompt and always prepend in the front, then not friendly for streaming applications
% we propose 
% contribution: demonstrate the unique design and shown advanatges  by experiment and analysis on 
%the same sequence length 

%TODO

In this work, we propose an alternative {\em modular} and {\em multitask} SpeechLLM design that is both {\em streamable} and   {\em efficient}. 
%in the sense that the extra LLM computation from introducing speech modality does not grow with either speech feature lengths or LLM sizes.
Our main contributions are summarized as follows:
\begin{itemize}
    \item To the best of our knowledge, this is the first open SpeechLLM solution that enables {\em streaming} and  {\em multitask at scale} (beyond ASR) at the same time. Moreover, the solution is end-to-end optimizable and allows LLM knowledge transfer to speech.
    \item Propose a different backbone architecture from the popular
    Speech-LLaMA variants that is based on cross-attention and read-write policy. The novel backbone unifies the offline and streaming modes and achieves state-of-the-art on several large-scale and multitask  speech-to-text benchmarks (ASR, AST, SQA, DynamicSuperb), with {\em lower training/inference cost}.
    %is efficient and
    
    %scalable and can be the backbone design for next generation SpeechLLMs.
    %The proposed method
    
    %This method is more efficient than previous GPT-style models like Speech-LLaMA~\cite{wu2023decoder,fathullah2023prompting,chen2024salm} and  has much less additional parameters than 
    %T5-style~\cite{2020t5} models like Flamingo~\cite{alayrac2022flamingo,kong2024audio}.
    % build and will release a multitask large scale  model that can do speech and audio understanding , SOTA on ..., strong on spoken QA ...
    
\end{itemize}

%ADD DETAIL: how many GPUs is used to train the model in a single day
%ADD DETAIL: how many GPUs is used to train the model in a single day
Moreover, we scale the model to 87K hours of speech data and the training can finish in one day. 
We will release the code and checkpoints of this multitask speech foundational model to promote offline and streaming SpeechLLM research.

\begin{figure}[]
    \centering
    \includegraphics[width=0.85\linewidth]{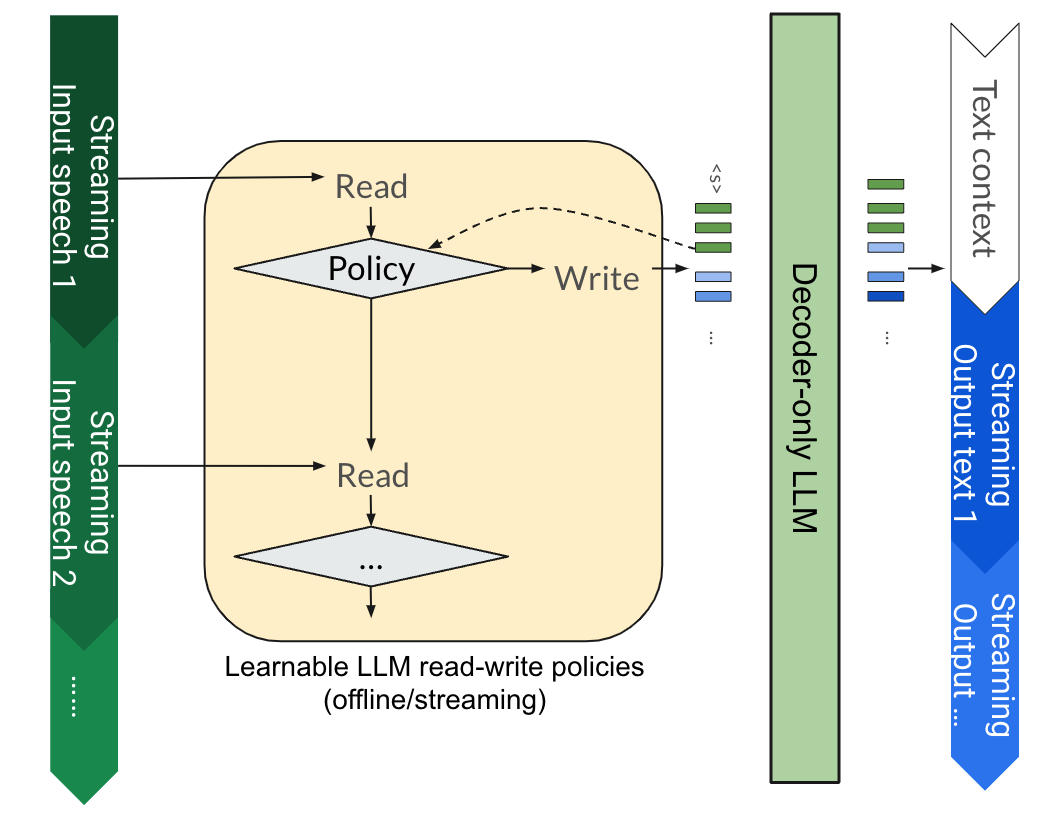}
    \vspace{-1em}
    \caption{Reformulate {\em streamable SpeechLLM} as a read-write policy problem previously used in simultaneous translation, i.e. 
    {\em the LLM agents should start replying whenever they think they have gotten enough information}.
    }
    \label{fig:framework}
    \vspace{-0.5em}
    % https://docs.google.com/presentation/d/1BHpzmTGq_RHKqh9_TNAq_pqxF1yvJcUa35kgsy2EtdI/edit#slide=id.g274b082b3fd_0_9
\end{figure}

\begin{figure*}[!tbh]
    \centering
    \includegraphics[width=0.9\linewidth]{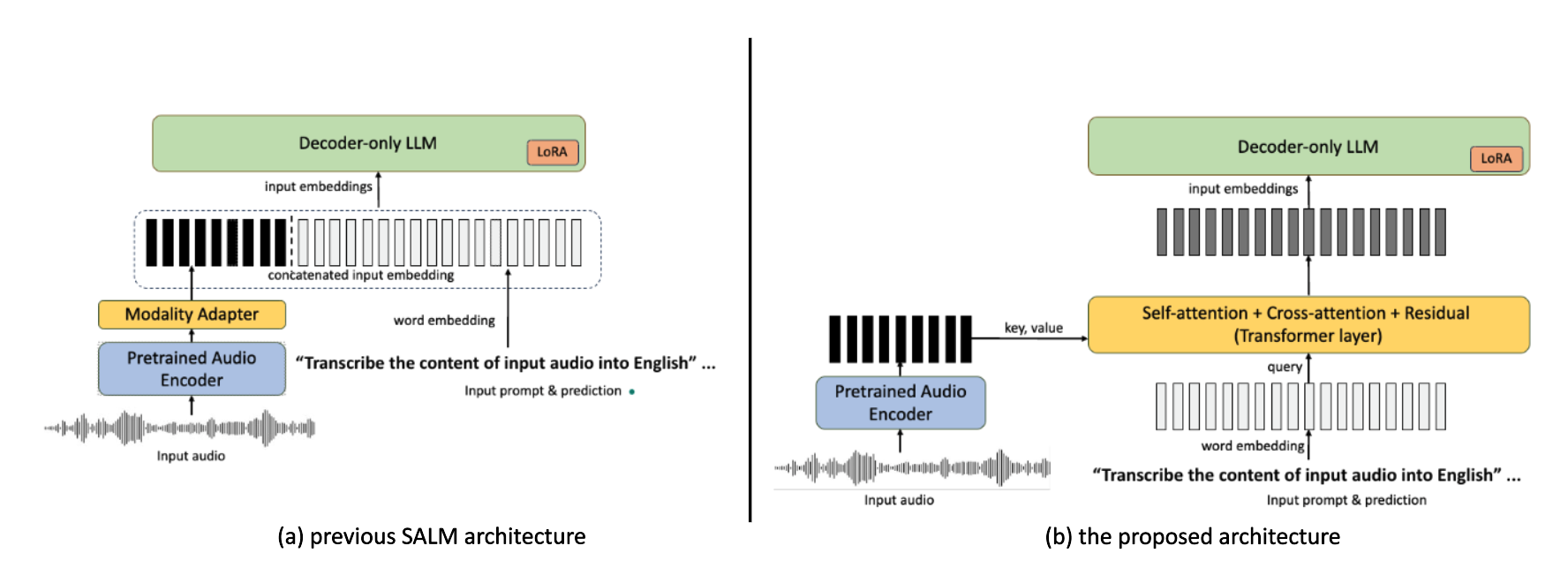}
    \vspace{-1.5em}
    \caption{(a)  SALM architecture~\cite{chen2024salm} (as an example of Speech-LLaMA). (b) the proposed \model architecture, which uses cross-attention to extract task-relevant features using the text as queries and speech features as keys/values. Compared with the former, the proposed model has lower runtime complexity, while achieving the state-of-the-art performance on multiple tasks.}
    \label{fig:model}
    \vspace{-0.5em}
    % https://nvidia-my.sharepoint.com/:p:/p/heh/EdXNOYspax5Em7rU-J7ChYABO_R-8_C6SvTowNA0N1xSfg?email=zhehuaic%40nvidia.com&e=Cm6BEJ
\end{figure*}

\section{Related work}

\subsection{Speech Foundational Model}
Motivated by the success of foundation models in NLP~\cite{brown2020gpt,team2023gemini}, recent speech foundational model research has been shifted towards developing universal pretrained models capable of handling multilingual speech and audio tasks. Recent advances include but are not limited to: i) Large scale multilingual self-supervised learning and semi-supervised learning  to leverage unlabled speech, e.g. XLSR~\cite{conneau2020unsupervised} and USM~\cite{zhang2023google}. ii) Large scale multitask supervised training, e.g. Whisper variants~\cite{radford2022robust,peng2024owsm} and SeamlessM4T~\cite{barrault2023seamless}. 
% why to build speech foundational model
% recent advances:
% self supervised and semi supervised learning; XLSR, USM
% large scale supervised training, whisper, owsm
% other tasks: SUPERB
iii) %\subsection{Audio Representation Learning}
More powerful and  efficient speech encoder design, e.g. 
%Speech models typically replace transformer with a modified architecture that is more suitable for speech modality. 
Conformer and variants~\cite{gulati2020conformer,rekesh2023fast}. 
%introduces convolutional modules to improve local context sensitivity. 
%FastConformer~\cite{} further modifies the architecture and increases the downsampling factor to 8, achieving 2.8x speedup without loss of modeling power.
iv) Multilingual multitask speech benchmarks, e.g. XTREME~\cite{conneau2022xtreme} and ML-SUPERB~\cite{shi2023ml}.

\subsection{SpeechLLM}
\vspace{-0.5em}
Recently, researchers started to look at 
combining pretrained speech models with  large language models to form a new type of speech foundational model, {\em SpeechLLM}, which can be  categorized through speech embedding types. SpeechGPT~\cite{zhang2023speechgpt} and AudioPaLM~\cite{rubenstein2023audiopalm}  quantize speech into discrete tokens, so speech tokens and text tokens can be combined into a single vocabulary. 
%making  it easy to finetune the LLM with speech data. 
Nevertheless, this approach is limited by the quality and diversity of the speech quantization model especially for STT tasks.   % and is not scalable to new quantization models. 
%When better quantization models are available, the SpeechLLM needs to be finetuned completely, since the vocabulary of speech tokens might have changed significantly in the new model.
%
Another more popular method is to feed  speech into pretrained speech encoder, where the speech features are then projected into the word embedding space of the LLM~\cite{chen2024salm,kong2024audio,qwen2023,tang2023salmonn,gong2023ltu-as}. Our work belongs to this category.

After extracting speech features, 
the previous architectures to provide speech information to LLMs
can be further categorized into two branches: 
i) {\em GPT-style} models~\cite{chen2024salm,tang2023salmonn,qwen2023} directly concatenate the speech prompts with text prompts and use the combined sequence as input to the LLM like a decoder-only GPT model; ii) 
Flamingo and its extension~\cite{kong2024audio,alayrac2022flamingo,radhakrishnan2023whispering} are another branch of works, where 
cross-modal cross-attention is added to each layer/block of the pretrained GPT-based LLM with shared text query. %, such that the text tokens can access speech information at every layer of the LLM. 
The resultant architecture is similar to the {\em T5 architecture}~\cite{2020t5}. 

%On one hand, GPT-style models have much less trainable parameters than T5-style models, but at the cost of having to extract good enough speech features before input to the LLM. Efficiency is another potential problem which will be elaborated in Section~\ref{sec:efficiency}. On the other hand, T5-style models have the benefit of directly accessing speech features at each layer of the LLM, making it easier to extract task relevant information, but the cost of introducing parameters of all cross-attention layers is large.

Some recent works look at introducing cross-attention layers into the first branch of methods, but with different motivations and resultant designs from our work. E.g. \cite{yu2023connecting,hussain2023m} introduces cross attention before the above concatenation 
to bridge the gap between the speech encoder and LLM, whose computation is the same or more.

\subsection{Streaming Speech Models}
%TODO(\red{Nithin}): help 2-3 sentences review of ASR streaming methods

In streaming ASR, Efforts have been made to enable streaming into Transformer~\cite{zhang2020transformer,noroozi2023stateful}. These methods either use limited context with offline models  or train models in a streaming manner. 
%These encoders utilize causal convolutions in the pre-encoder layers with various left and right contexts depending on the suitable streaming conditions during inference. This simulates streaming inference and uses a cache mechanism to enhance performance with additional context. 
The methods usually build on top of Transducer~\cite{graves2012sequence} which rarely benefits from pretrained LLM.

In simultaneous speech translation, researchers have been looking at  fixed and adaptive read-write policies. Wait-k~\cite{ma2020simulmt} and variants are the most longstanding fixed policy. Typical adaptive policies try to learn the policy with a bespoke model from the training data, e.g. EMMA~\cite{ma2023efficient} and ITST~\cite{zhang2022information}.

\section{Why Streaming SpeechLLM is hard}
\label{sec:prev_stream}
\vspace{-0.5em}
While making SpeechLLMs  operate in real-time and proactively respond is important for human-machine interface, the topic is under-explored besides some very recent works. 

As SpeechLLM usually starts offline text prediction after it accepts a complete speech segment as the speech prompt, these streaming works propose to update the prompt format to 
take interleaved block-wise speech features and text~\cite{seide2024speech,tsunoo2024decoderonly}.  \cite{agostinelli2023simul,koshkin2024transllama} also belongs to this category except focusing on text machine translation with an ASR model as cascaded speech frontend. There are several limitations in this line of approaches:
   i) \textbf{Prompt format mismatch} between text pretrain and offline/streaming modes of SpeechLLM.
   The updated interleaved format with injected {\scriptsize BLANK}\cite{seide2024speech} or {\scriptsize WAIT}\cite{koshkin2024transllama} tokens  circumvents the LLM textual knowledge transfer to speech. Previous research cannot show extra wins from leveraging pretrained LLMs  and  didn't demonstrate abilities  beyond  ASR~\cite{seide2024speech,tsunoo2024decoderonly}. 
ii)
 \textbf{Not end-to-end optimizable} blocked by the introduced alignment stage. This alignment preprocess design not only results in hardness in generalizing to multitask, e.g. AST, SQA, but also is bounded by the alignment errors in e.g. CTC \cite{seide2024speech} and translation \cite{koshkin2024transllama}, especially considering certain word can cross speech segments.
iii) 
\textbf{Higher inference cost},  stems from either longer text prediction length in the new prompt format~\cite{seide2024speech} or requiring beam search~\cite{wang2023simultaneous}.

A new streamable SpeechLLM solution will be proposed  which keeps the same LLM prompt format and unifies the learning framework of offline and streaming modes. Several unique advantages that will be demonstrated: i) \textbf{LLM knowledge transfer} ii) \textbf{multitask support} iii)\textbf{ end-to-end optimizable} iv) \textbf{lower training/inference cost}.  Lastly, we believe this is the first open-source streamable SpeechLLM solution.

\section{BESTOW: a Streamable SpeechLLM}
\vspace{-0.5em}
%One design principle of the proposed method is to connect the pretrained speech encoder to the LLMs in a modular way so that the LLM backbone is unchanged and textual knowledge can be transferred. 

%The proposed method is in line with Speech-LLaMA~\cite{wu2023decoder,fathullah2023prompting,chen2024salm} in the sense that only the input of LLM is updated while the rest is unchanged or finetuned with LoRA~\cite{hu2021lora}. This modular design  ensures the textual knowledge and in-context learning ability transfer.
%
%Nevertheless, we propose a new mechanism to cross-attend speech and text prompts in the input side, which shows unique advantages in streaming, efficiency, and scalability.

\subsection{Unified offline-streaming framework}
The goal is to design a unified framework for offline and streaming SpeechLLM so as to maximize the LLM knowledge transfer from pretrain and instruction tuning. Offline and streaming modes should share most of the architectures and end-to-end optimizable. %, different from previous works in Section~\ref{sec:prev_stream}  

As Figure~\ref{fig:framework}, we propose to formulate the streaming problem of SpeechLLM as the read-write policy problem previously defined in simultaneous speech translation~\cite{ma2020simulmt}. % where offline SpeechLLM is generalized through offline policy.
At every LLM step, the model  decides whether to wait for more speech incoming features (READ) or to predict a target word (WRITE).
The solution still keeps the prompt format of LLM unchanged and experiment results will show its benefit on LLM knowledge transfer. 
The prerequisite of this solution is to decouple  READ and WRITE operations  from LLM and make it modeled by a standalone module, the cross-attention feature extractor proposed in Section~\ref{sec:xattn}. 
Empirical result will show this module is on par with Speech-LLaMA architecture. % in apple-to-apple setting. % and performs the best among open SpeechLLM.
After that, streaming is straightforward and end-to-end optimizable, discussed in Section~\ref{sec:stream}.
Lastly we  discuss additional efficiency benefit of this architecture in Section~\ref{sec:efficiency}.

\subsection{Introduce speech modality with cross-attention}
\label{sec:xattn}
\vspace{-0.5em}
% original version:
%The main novelty of the model lies in how s interact with speech features. Different from previous works~\cite{chen2024salm,tang2023salmonn} that simply concatenate speech features with s as input to the LLM, we use a cross-attention transformer module with residual connections to let s attend to different regions of speech features to extract features that are more relevant to the given task. As illustrated in Figure~\ref{fig:model}(b), the speech features extracted from the speech encoder serve as the keys and values for the cross-attention Transformer module, while the input text is embeded by the LLMs token embedding matrix and then serves as the queries to the cross-attention Transformer. The cross-attention Transformer has multiple layers of interleaving cross-attention and self-attention as in the original TransformerDecoder~\cite{vaswani2017attention}. To allow propagating more information directly from the speech features, we add a residual connection that adds speech features directly to the output of the cross-attention Transformer, and the combined embeddings serve as the final input to the pretrained LLM. 

We propose a new mechanism on how text prompts interact with speech prompts
%to cross-attend speech and text prompts 
in the LLM input side, which will show unique advantages in streaming and efficiency. 
%
%The main novelty of the model lies in . 
Different from majority~\cite{wu2023decoder,fathullah2023prompting,chen2024salm,tang2023salmonn} of previous works that simply concatenate speech prompts with text prompts as input to the LLM (Figure~\ref{fig:model}(a)), 
we inject a trainable 
 transformer-like self-attention  and cross-attention layers  before feeding into LLMs 
to let text prompt attend to different regions of speech prompt to extract features that are more relevant to the current LLM step
as illustrated in Figure~\ref{fig:model}(b).

With this new design, the speech features extracted from the speech encoder serve as the keys and values for the cross-attention mechanism, while the input text (text prompt and previous tokens) is firstly embedded by the LLMs input embedding matrix and later used as the queries. 
% add self attention to model history
To make the query considering both the current step and the context, we further inject causal self-attention layers between the LLM input embeddings and the cross-attention layers. In the ablation study section, we also consider an alternative design of RNNs for the same goal. 
%To match the speech and text modeling space, a feed-forward network is further added to the cross-attention outputs.
%To allow propagating more information directly from the speech features, we add a residual connection that adds speech features directly to the output of the cross-attention Transformer, 
To preserve the original textual knowledge,
we include a residual connection that adds text prompts directly to the output 
and the combined speech and text embeddings serve as the final input to the pretrained LLM. 
The resultant design is essentially one cross-attention transformer layer proposed in \cite{vaswani2017attention} which can be repeated for $X$ times. We empirically found in the ablation study section that $X=2$ is sufficient. % where all the LLM blocks can be frozen with LoRA~\cite{hu2021lora} finetuning. 
Besides, the residual design above  ensures the model can fall back to the  original textual LLM by learning to ignore the cross-attention outputs in non-speech steps.
%The cross-attention Transformer has multiple layers of interleaving cross-attention and self-attention as in the original TransformerDecoder~\cite{vaswani2017attention}.

%The proposed model can be trained using next-token-prediction as in training other GPT models~\cite{brown2020gpt}. During inference, after feeding all text prompt to the cross-attention Transformer, the LLM starts to genrate output tokens, where each predicted token is fed back to the cross-attention Transformer so that it also attends to the speech features like the input text prompt.

The proposed model can be trained using next-token-prediction as in training other GPT models~\cite{brown2020gpt}. 
During inference, 
LLM can take text prompts and generate 
output tokens in a step-by-step fashion.
%after feeding all text prompt to the cross-attention Transformer, the LLM starts to genrate 
The only difference is that  each predicted token is fed back to both the input of LLM and  the cross-attention module above. % so that it also attends to the speech features like the input text prompt.
% describe step by step prediction

% compare to ASR streaming SOTA
% include computational aware latency 
% include latency and accuracy tradeoff 
% be careful about describing how to do streaming; ask someone working on streaming to proofread  

\subsection{From Offline to Streaming }
\label{sec:stream}
\vspace{-0.5em}
%TODO(\red{Nithin}): can you help add the details of our current cache-aware unidirectional encoder (also include the dataset information here for now; we will later move some of them to 4.1)

With the above cross-attention speech feature extractor, the speech context length required to make prediction at each decoder step is independent with LLM backbone and completely decided by the cross-attention module with a read-write policy in Figure~\ref{fig:framework}. 
This characteristic enables the streaming design in simply two steps.
%We believe this characteristic will be very useful for many streaming applications in speech by augmenting the design with monotonic cross attention and chunk-wise encoder layers~\cite{raffel2017online,ma2019monotonic}, which will be our future topic.  

The first step is to design a read-write policy for handling the streaming speech input. 
Our framework converts streamable SpeechLLM to 
a similar read-write problem as simultaneous translation where previous research in fixed and adaptive policies can be reused.
We will take the most popular
and longstanding fixed policy, {\em wait-k}~\cite{ma2020simulmt}, as an example in the following while integrating more adaptive policies~\cite{ma2023efficient} will be  interesting future topics. 
We first decide a fixed pre-decision ratio of $L$ which represents a step size of $(L*P*10\tt{ms})$ where $P$ is the downsampling ratio of the speech encoder.
After taking text context prompt without cross-attention,  LLM starts to predict the first subword unit by cross-attending to the first $(K*L)$ speech embedding steps. After that, LLM predict one next subword for every $L$ incoming  speech embeddings in a streaming fashion.

The second step is to make the speech encoder work in streaming mode. This can be done  by two approaches: i) keep the bidirectional encoder in the inference time, recompute all the available encoder states after getting each new speech block, and provide that to the above read-write policy (denoted as {\em BESTOW-S-bidi}) ii) retrain the model with a unidirectional encoder (denoted as {\em BESTOW-S-unid}). 
For the former, we introduce a fixed right context of 13 frames in the inference time to compensate the training/inference mismatch.
For the latter, we adapt the cache aware streaming model\cite{noroozi2023stateful} to update the FastConformer pre-encoder layers. We utilize causal convolutions with the following left and right context windows in frames: [[70,13],[70,6],[70,1],[70,0]],
which can be chosen from in the inference time.
%These range from 70 frames (5 sec) for left context to 0 frames (10msec) for right context. 
%This makes it suitable for streaming scenarios in various contexts as verified by previous ASR research. 

We initialize the streamable SpeechLLM ({\em BESTOW-S}) from the offline SpeechLLM ({\em BESTOW}) and continue training on the same data.
To improve the  generalization, we train BESTOW-S with a random $K$ range so that in  inference, any $K$ in the range can be used to allow latency-quality tradeoff.

\subsection{BESTOW v.s. GPT-style and T5-style}
\label{sec:efficiency}
Besides the  streamable capability above, compared with GPT-style SpeechLLMs (or Speech-LLaMA) ~\cite{chen2024salm,tang2023salmonn,gong2023ltu-as}, the proposed architecture is also computationally more efficient. Let $L_t$ and $L_a$ denotes the lengths of text tokens and speech features respectively, then computational complexity of GPT-style models is to the order of $(L_t+L_a)^2 = L_t^2+L_a^2 + 2L_aL_t$, due to the quadratic complexity for self-attention mechanism. Meanwhile, with our cross-attention between text and speech prompts, we are able to reduce the complexity to $L_tL_a + L_t^2$. Given that the length of text tokens is usually much shorter than that of speech features (in which case $L_a \gg L_t$), in theory we can enjoy a speed up of $L_a$ times, which means the longer the speech the greater the speedup. 
This speedup is crucial especially considering very wide and deep LMs where the computation from speech encoder and cross attention can be omitted from  LLM forward and backward computation.  
To compensate for that, all previous research has to introduce significant downsampling on the speech features to reduce the feature length, which potentially incurs information loss.
%One might argue that, the GPT-style SpeechLLMs can attend to the speech features at each layer of the LLM, 
%while our model can only do that before LLM. %, which might lose some performance. 
%while our model can only do that before LLM, which might lose some performance. 
%However, our experiments will show that the proposed model is able to achieve better or the same performance as GPT-style Speech-LLaMA.

Although the T5-style models like Flamingo~\cite{alayrac2022flamingo,kong2024audio} and Whisper-LLaMA~\cite{radhakrishnan2023whispering}  also add cross-attention on pretrained GPT-based LLMs, the cross-attention is added to each layer of the LLM, which introduces a large number of parameters. In contrast, we show  with only two layers of cross-attention before feeding into LLMs, we can achieve state-of-the-art performance in ASR and AST. 
% discuss LoRA finetune
Moreover, the modality-specific design in the proposed method is only introduced to the input of LLMs, which allows isolating the LLM parameter updates to  parameter-efficient-finetune methods  like LoRA~\cite{hu2021lora} as demonstrated in the experiment.

\begin{table*}[!tbh]
\centering
\caption{Compare BESTOW, as a {\em Multitask} and {\em Streamable} SpeechLLM, with other multitask SpeechLLMs. For ASR, we use Librispeech test-other ({\em LS})\cite{panayotov2015librispeech} and Gigaspeech ({\em Giga})\cite{chen2021gigaspeech}  and report {\em non-computation-aware LAAL} and {\em WER} as latency and quality metrics. For AST, we use CoVoST~\cite{wang2020covost}  and report {\em LAAL} and {\em BLEU}. We  report numbers on speech synthesized MSMACRO ({\em SQA})\cite{data_gen_2024} and  DynamicSuperb\cite{dynamicsuperb_leaderboard} (follow \cite{huang2023dynamic} to report 6 categories)  where {\em ROUGE-L} and {\em Accuracy} are reported.  }
\label{tab:speechllm-compare}
\resizebox{\textwidth}{!}{
\begin{tabular}{l|c|cc|c|cccc|c|cccccc}
\hline
Model           & \multicolumn{3}{c|}{ASR $ \downarrow$} & \multicolumn{5}{c|}{AST $\uparrow$} & SQA $\uparrow$ &  \multicolumn{6}{c}{DynamicSuperb (unseen) $\uparrow$}                          \\
            & {\em LAAL} $\downarrow$& LS        & Giga       &{\em LAAL} $\downarrow$& en-de &de-en  & es-en  & fr-en &          &CNT &SPK&SEM&DEG&PRL&AUD             \\

\hline\hline
\multicolumn{9}{l}{{\em Offline}  multitask SpeechLLM}                                                                              \\
\hline
QWEN-audio\cite{qwen2023}  & N/A        & 4.2       & 10.2    & N/A   &25,1  & 33.9   & 39.7   & 38.5  & 39.7   &  \multicolumn{6}{c}{N/A}                         \\
SALMONN\cite{tang2023salmonn}     & N/A        & 4.9       & 10.0    & N/A   &18.6 & N/A    & N/A    & N/A   & 48.8    & \multicolumn{6}{c}{N/A}                         \\
Whisper-LLM\cite{huang2023dynamic} & N/A        & N/A       & N/A  & N/A   &N/A    & N/A    & N/A    & N/A   & N/A  & 8.7 & 60.6 & 20.9 & 59.0 & 6.6 & 15.9  \\
BESTOW      & N/A        & \textbf{3.2}      &\textbf{9.9}     & N/A &39.0     & \textbf{39.0}   & \textbf{41.1 }  & \textbf{41.0}  & \textbf{59.8} & 
100.0&81.0 &75.0&43.5&46.0&2.0\\
\hline\hline
\multicolumn{9}{l}{{\em Streamable} multitask SpeechLLM}                                                                           \\
%./results/test.ast.crossbmg4eghel_lhmerge_oci_FC-GPT_llama_tiny_canaryset_b6s4kf-sunolong_noCC_langtemp0.5_dsettemp0.5_lr1e-4wd1e-3_CosineAnnealing_warmup2500_minlr1e-6_gbs4096_mbs16_ep200/off_test_covost_en_de_inputs_preds_labels.jsonl
\hline
%dynamic superb https://docs.google.com/document/d/1omTiogocsFXo_pO-iWDrYUe_b5ndzt56sMx1OGc5QkM/edit 
% 5 8
%BESTOW-S-bidi   & 2.9        & 3.4         & 10.5      & 4.9   &38.6  & 37.9   & 40.4     & 40.5  & 52.3    &  \multicolumn{5}{c}{N/A}  \\ 
% 10 4
BESTOW-S-bidi   & 3.0        & 3.5         & 10.4      & \textbf{3.9}   & \textbf{39.2}  & 37.9   & 40.1     & 40.3  & 52.3    &  94.0 & 59.5 & 76.5 & 39.0 & 47.0 & 1.0 \\ 
BESTOW-S-unid    & \textbf{2.9}        & 4.3         & 10.8     & \textbf{3.9}  &36.6    & 34.9   & 37.7     & 37.6  & 47.7    &  97.5&79.5&76.5&76.5&39.0&2.5  \\ \hline
\end{tabular}
}
\vspace{-1em}
\end{table*}
%'/lustre/fsw/portfolios/llmservice/users/zhehuaic/results/canary-v0_speechllm/crossbmg4eghel_mored_lhmain3cross_oci_FC-GPT_llama_tiny_canaryset_b6s4kf-sunolong_noCC_langtemp0.5_dsettemp0.5_lr1e-4wd1e-3_CosineAnnealing_warmup2500_minlr1e-6_gbs4096_mbs16_ep200/crossbmg4eghel_mored_lhmain3cross_oci_FC-GPT_llama_tiny_canaryset_b6s4kf-sunolong_noCC_langtemp0.5_dsettemp0.5_lr1e-4wd1e-3_CosineAnnealing_warmup2500_minlr1e-6_gbs4096_mbs16_ep200/checkpoints/megatron_audio_gpt_peft_tuning--validation_bleu=50.129-step=60000-epoch=2.ckpt'
%https://wandb.ai/zhehuaic/canary-v0_speechllm/reports/w-and-w-o-qa-and-dynamic-superb--Vmlldzo3MzcxMzQ5
%https://docs.google.com/document/d/1iFR9wnjTji4j4caexGc21hcJc-qbm1_k-ycSBfSd488/edit

\section{Experiments}

\subsection{Datasets and Settings}

We  include  31K hours of public data speech as~\cite{canary-1b}  and 54K hours of extra in-house data for speech recognition (ASR) and speech translation (AST), 
%, which consists of 31k hours of public data, 
%20k hours collected by Suno, 
%and 34k hrs of in-house data. Among the 85K hours of ASR/AST data, 
which includes 67.4K hours in English, 6.1K hours in German, 6.6K hours in Spanish, and 5.1K hours in French.
In order to support multitask speech and audio understanding, we further add 2K hours from the speech synthesized version (released in \cite{data_gen_2024}) of MSMACRO text QA~\cite{bajaj2016msmacro} and the training set of DynamicSuperb~\cite{huang2023dynamic}.

We evaluate ASR and AST performance through WER and BLEU on public benchmarks elaborated in the next section. SpeechQA (SQA) and general speech understanding ability are assessed through the test splits of the above  datasets respectively. 
%To evaluate the in-context learning ability of the proposed architecture, we conduct contextual word boosting through  prompting LLM with a list of word candidates in the text format as~\cite{chen2024salm}.
% TODO(zhehuai): cite Andrei's paper which will talk about the GTC dataset release
%We follow the NVIDIA GTC talks setup\cite{fast_biasing}, which
%includes a large number of different acronyms, product names, and technical terms~\footnote{Examples: {\em NVIDIA, GPU, Omniverse, Geforce, kubernetes,} etc.} and  often results in poor recognition accuracy for ASR systems.
%We include 32 keywords in this study and report F-score of with and without in-context learning.
We also report  {\em non-computation-aware LAAL}\cite{papi2022over}  as the latency metric of the streamable {\em BESTOW-S}.

We implement the model with PyTorch using  NeMo Tookit~\cite{kuchaiev2019nemo}, and the model is trained on 128 A100 (80G) GPUs for 60k steps in one day, with a batch duration of 360 sec per GPU. The speech encoder is initialized from the Canary-1B model~\cite{canary-1b}, while the LLM is initialized from the TinyLlama-1.1B-chat model~\cite{zhang2024tinyllama}. We use a 2-layer cross-attention Transformer module, where effect of its layers is studied in Section~\ref{sec:exp-offline} with separate experiments on LibriSpeech~\cite{panayotov2015librispeech}. We train all parameters in the model by default, which is about 1.8 billion. LoRA~\cite{hu2021lora} with frozen LLM backbone is also explored in an ablation study (1024 dimensions). 
We use distributed fused Adam optimizer and cosine annealing, with learning rate $1e-4$ and weight decay of $1e-3$. Gradient clipping of $1.0$ is  applied. Code, configs and checkpoints will be publicly available.

 %TODO: add streaming details.
We follow Section~\ref{sec:stream}
 to turn above offline {\em BESTOW} to streamable {\em BESTOW-S}. We set $L=4$, $P=8$, $K=10$ by default for all tasks except $K=6$ in ASR task as it requires less speech context. $K$ is sampled from $3$ to $12$ in training. Varying $L$ to result in different step size is studied in Section~\ref{sec:lat-trade}. 
%TODO: shorten the following
To stabilize the training of {\em BESTOW-S-unid}, we first pretrain the FastConformer cache aware encoders~\cite{noroozi2023stateful} using wav2vec2 on the librilight~\cite{librilight}. 
In the inference time, we always use 13 frames as the right context of the speech encoder of   {\em BESTOW-S-unid} and  {\em BESTOW-S-bidi}.
%We trained the model for 130,000 steps with a global batch size of 360 seconds. 
%Then, we fine-tune the model on Canary data using a transformer decoder for 100,000 steps.

We compare the proposed cross attention based SpeechLLM architecture with 
SALM\cite{chen2024salm} trained on the same data in Section~\ref{sec:exp-offline},  which is an open-source speech language model  conditioning a LLM on speech and text prompts to predict textual outputs for different speech tasks. 
%The main difference is that 
%instead of drawing a cross attention between the speech prompts and the original textual LLM inputs, 
SALM simply prepends speech prompts to the text prompts as the LLM inputs which is the typical design of Speech-LLaMA~\cite{wu2023decoder,fathullah2023prompting}.
%To make sure parity in this setting, 
%we freeze LLM backbone and train with the same 256 dimensional LoRAs.  

\begin{table*}[!tbh]
\centering
\caption{Comparison with state-of-the-art models on speech recognition (ASR) and speech-to-text translation (AST). For ASR, we use MCV-16.1~\cite{commonvoice:2020} test sets, and process both the predictions and groundtruth using WhisperNormalizer~\cite{radford2022robust} to report {\em WER}. We also report average WER on large-scale Open ASR Leaderboard~({\em HF-Lead})\cite{open-asr-leaderboard}. For AST, we use FLEURS~\cite{conneau2023fleurs} test sets and their native annotations with punctuation and capitalization to report {\em BLEU}.}
\label{tab:asrast}
\resizebox{\textwidth}{!}{
\begin{tabular}{l|l|crrrr|rrrrrr}
\hline
\multicolumn{1}{c|}{\multirow{2}{*}{Model}} &Data& \multicolumn{5}{c|}{ASR (WER $\downarrow$)} & \multicolumn{6}{c}{AST (BLEU $\uparrow$)} \\ 
\multicolumn{1}{c|}{} & hour & \multicolumn{1}{c|}{HF-Lead} & \multicolumn{1}{c|}{En} & \multicolumn{1}{c|}{De} & \multicolumn{1}{c|}{Es} & \multicolumn{1}{c|}{Fr} & \multicolumn{1}{c|}{En$\rightarrow$De} & \multicolumn{1}{c|}{En$\rightarrow$Es} & \multicolumn{1}{c|}{En$\rightarrow$Fr} & \multicolumn{1}{c|}{De$\rightarrow$En} & \multicolumn{1}{c|}{Es$\rightarrow$En} & \multicolumn{1}{c}{Fr$\rightarrow$En} \\ \hline \hline
SeamlessM4T-medium (1.2B) & 4M& \multicolumn{1}{c|}{N/A} & \multicolumn{1}{r|}{10.25} & \multicolumn{1}{r|}{9.32} & \multicolumn{1}{r|}{7.25} & 11.07 & \multicolumn{1}{r|}{28.30} & \multicolumn{1}{r|}{21.05} & \multicolumn{1}{r|}{37.36} & \multicolumn{1}{r|}{33.39} & \multicolumn{1}{r|}{21.68} & 31.02 \\ \hline
SeamlessM4T-large-v2 (2.3B) & 4M& \multicolumn{1}{c|}{N/A} & \multicolumn{1}{r|}{7.47} & \multicolumn{1}{r|}{5.82} & \multicolumn{1}{r|}{4.82} & 7.75 & \multicolumn{1}{r|}{\textbf{33.17}} & \multicolumn{1}{r|}{\textbf{23.72}} & \multicolumn{1}{r|}{\textbf{43.05}} & \multicolumn{1}{r|}{\textbf{37.06}} & \multicolumn{1}{r|}{\textbf{25.41}} & 30.94 \\ \hline
Whisper-large-v3 (1.5B) & 5M& \multicolumn{1}{c|}{7.16} & \multicolumn{1}{r|}{9.92} & \multicolumn{1}{r|}{6.17} & \multicolumn{1}{r|}{4.94} & 11.18 & \multicolumn{1}{r|}{N/A} & \multicolumn{1}{r|}{N/A} & \multicolumn{1}{r|}{N/A} & \multicolumn{1}{r|}{33.40} & \multicolumn{1}{r|}{22.70} & 33.70 \\ \hline
Canary-1b (1B) & 85K&\multicolumn{1}{c|}{6.67} &  \multicolumn{1}{r|}{7.97} & \multicolumn{1}{r|}{4.61} & \multicolumn{1}{r|}{3.99} & 6.53 & \multicolumn{1}{r|}{32.15} & \multicolumn{1}{r|}{22.66} & \multicolumn{1}{r|}{40.77} & \multicolumn{1}{r|}{33.99} & \multicolumn{1}{r|}{21.80} & 30.95 \\ \hline
\hline 
\model (1.8B) &87K &\multicolumn{1}{c|}{\textbf{6.50}} &  \multicolumn{1}{r|}{\textbf{7.31}} & \multicolumn{1}{r|}{\textbf{4.16}} & \multicolumn{1}{r|}{\textbf{3.77}} & \textbf{6.18} & \multicolumn{1}{r|}{31.98} & \multicolumn{1}{r|}{23.08} & \multicolumn{1}{r|}{41.90} & \multicolumn{1}{r|}{35.75} & \multicolumn{1}{r|}{23.86} & \textbf{35.10} \\ \hline
\end{tabular}}
\vspace{-1em}
\end{table*}

\subsection{BESTOW-S enables streaming and multitask}

Table~\ref{tab:speechllm-compare} compares the proposed BESTOW, as a {\em Multitask} and {\em Streamable} SpeechLLM, with other SpeechLLMs. 
To our best knowledge,
BESTOW-S is the only open SpeechLLM that supports both streaming and multitask
\footnote{We acknowledge that previous streaming works~\cite{seide2024speech} and \cite{tsunoo2024decoderonly} report 7.4 and 7.9 on Librispeech of ASR task (our number is 3.3) but not included   because of  only supporting ASR and  not using or benefiting from LLM.}.
Both BESTOW and BESTOW-S significantly outperform other models on ASR and AST tasks. The BESTOW models can also support SQA and unseen tasks from DynamicSuperb and perform reasonably well even in streaming mode (BESTOW-S) with a 3.9 seconds LAAL lagging.
As a contrast for the SQA task~\footnote{Given no common SQA benchmark, while other SpeechLLM works do not train on the SQA dataset~\cite{data_gen_2024} we used, it may not be  fair to compare.}, we  built a strong cascaded baseline  by first transcribing the speech with  110M NGC ASR pretrained Fast Conformer-large, followed by feeding hypotheses to a LoRA finetuned TinyLlama 1B LM with the textual MSMARCO dataset which leads to 56.2 ROUGE-L. BESTOW performs decently (59.8 and 52.3) compared to the cascaded baseline.

In short, 
BESTOW-S enables streaming and multitask at the same time with competitive performance compared to offline BESTOW on each task. The latency is tunable and can be less on tasks like ASR which require less context. Section~\ref{sec:lat-trade} will  elaborate on the latency-quality tradeoff. 
BESTOW-S-bidi with bidirectional encoder and recomputation in the inference time is generally better than BESTOW-S-unid, consistent with previous research in SST~\cite{ma2023efficient,ma2020simulmt}. Optimizing the computation-aware lagging of the bidirectional encoder and closing the gap between bidirectional and unidirectional solutions are important next steps.

\subsection{Can BESTOW serve well as a  offline SpeechLLM?}
\label{sec:exp-offline}
In this section, we take a step back and confirm whether BESTOW as a offline/streaming unified architecture can perform well on offline scenario.  
We compare the offline BESTOW with the state-of-the-art ASR models on MCV-16.1~\cite{commonvoice:2020} and AST models on FLEURS~\cite{conneau2023fleurs} test sets, and show the results in Table~\ref{tab:asrast}.
We use the Whisper~\cite{radford2022robust}, SeamlessM4T~\cite{barrault2023seamless} and Canary~\cite{canary-1b} as baselines\footnote{Acknowledging strong results from  Gemini\cite{team2023gemini} and SLM\cite{wang2023slm}, we cannot compare to them due to missing per-language numbers.}, by using their official checkpoints and rerunning the models on the same test sets. All models use beam search with beam size 5.

BESTOW performs the best in the multidomain Open ASR Leaderboard~\cite{open-asr-leaderboard} and the four seen languages of multilingual ASR benchmark. Moreover, it achieves the first place of one AST language pair and  second places of another four pairs with 2\% of the training amount of SeamlessM4T-large-v2~\cite{barrault2023seamless}.
Notably, while BESTOW is built on top of the speech foundational model Canary-1b, it  significantly advances ASR and AST performances on all tasks in the last two rows.
This demonstrates that BESTOW architecture can {\em leverage LLM textual knowledge} to improve speech task performance and {\em minimize the data amount} to build good systems.

%In Table~\ref{tab:SpeechLLM}, we further compare BESTOW with recent SpeechLLM works on ASR (Librispeech-test-other~\cite{panayotov2015librispeech}, Gigaspeech~\cite{chen2021gigaspeech}) and AST (CoVoST~\cite{wang2020covost})  datasets. With smaller LLM backbone, BESTOW not only achieves the lowest WER on ASR datasets, but also significantly advances the AST performance on all the language pairs seen in the training.

%\subsection{Results on Speech and Audio Understanding}
%Besides ASR and AST, the proposed BESTOW can also perform general speech and audio understanding as shown by Table~\ref{tab:multitask}.  We leverage DynamicSuperb benchmark to evaluate on five representative tasks~\cite{dynamicsuperb_leaderboard} from DynamicSuperb~\cite{huang2023dynamic} and report average accuracy by categories. 
%We compare to SOTA systems from the DynamicSuperb paper~\cite{huang2023dynamic}.
%BESTOW performs well on Content, Semantics and Speaker dimensions and behind on Audio dimension, which is understandable as 99\% of the training data is non-audio speech data. Improving audio performance through pretrained encoder and training data will be our next step.

%Table~\ref{tab:qa} shows the result on speech question-and-answer ({\em speechQA}) task to  evaluate speech content understanding ability.
%

Table~\ref{tab:prependvs}  conducts another apple-to-apple comparison between the proposed cross-attention speech feature extractor ({\em X-Attn}) in Section~\ref{sec:xattn} and 
one of the most popular offline methods to connect speech models with LLM, Speech-LLaMA~\cite{wu2023decoder,fathullah2023prompting} types of works ({\em Prepend}).
The latter branch of methods share an architecture of prepending speech embeddings to the text  embeddings before feeding to a decoder-only LLM. We leverage
the open-source SALM implementation~\cite{chen2024salm} to build a model with the same training data and similar size for comparison, which uses
two Conformer layers with 4X subsampling on top of the pretrained speech encoder as modality adapter layers.  The speech encoder is always jointly trained.
{\em Prepend} performs similar as {\em X-Attn} on ASR and AST tasks while the proposed {\em X-Attn} shows clear speed and memory improvements as explained in Section~\ref{sec:efficiency}. We also contrast LoRA finetune with full finetune LLM backbone in the last row. The ASR and AST results are similar while in practice this LoRA based BESTOW model takes much longer time to converge (170k v.s. 60k steps). 

Table~\ref{tab:ablation} details  the values from each  introduced component in the cross-attention speech feature extractor. 
The speech encoder here is always initialized with a smaller 110M NGC ASR pretrained Fast Conformer-large %\footnote{\scriptsize\url{https://catalog.ngc.nvidia.com/orgs/nvidia/teams/nemo/models/stt_en_fastconformer_transducer_large}}
and  trained.
In this study, we found two layers of transformer-like self-attention ({\em Self-Attn}) and cross-attention ({\em X-Attn}) layers are sufficient to get the optimal performance.
Removing self-attention in the layer results in significant degradation. 
As described in Section~\ref{sec:xattn}, the self-attention is crucial to model the  history of previous text tokens so as to be the query for the latter cross-attention. 
To confirm this, we also tried an alternative design in the last  row by using two layers of RNNs to model the history, similar to LAS architecture for end-to-end ASR~\cite{chan2016listen}, which can effectively bring back the WER. 
%In the last row, we unfreeze the 1B LM backbone, which obtains the best result for this task.
%The resultant system performs similarly to the SALM baseline.

\begin{table}[]
\caption{Efficiency and Accuracy Comparison of Speech-LLM Architecture. The training speed and memory consumption are measured by steps-per-second and memory consumption percentage of a A100 (80G) GPU. ASR and AST results are reported on MCV-16.1 and FLEURS respectively.  }
\label{tab:prependvs}
\resizebox{\columnwidth}{!}{
\begin{tabular}{l|ll|llll}
\hline
& Speed $\uparrow$ & Mem. $\downarrow$ & \multicolumn{2}{c}{ASR $\downarrow$} & \multicolumn{2}{c}{AST $\uparrow$} \\
& (step/s) & (\%)  & en &de & en-de & de-en \\
\hline\hline
Prepend              & 0.50                                                                 & 69                                                            & 7.2                                                 & {4.1}                                                & \textbf{32.1}                                               & 36.0                                               \\
X-Attn      & 0.59                                                                 & \textbf{58}                                                           & 7.3                                              & {4.1}                                             & 32.0                                               & 35.8                                               \\
\  + LoRA & \textbf{0.60}                                                                    & N/A                                                               & \textbf{7.0}                                                 & {4.1}                                                & 31.6                                                   & \textbf{36.2}                                \\
\hline
\end{tabular}
}
\end{table}

%format of this table 
\begin{table}[]
\centering
\caption{Ablation study on the cross-attention architecture. A smaller encoder is trained while LLM is frozen with 256-dim LoRA. WER is reported on Librispeech testother split.}
\label{tab:ablation}
\resizebox{\columnwidth}{!}{
\begin{tabular}{l|l|r|r}
\hline
\multicolumn{1}{c|}{Arch.} & \multicolumn{1}{c|}{Details} & \multicolumn{1}{c|}{\begin{tabular}[c]{@{}c@{}}Trainable \\ parameters\end{tabular}} & \multicolumn{1}{c}{\begin{tabular}[c]{@{}c@{}}LS  \\ (WER)\end{tabular}} \\ \hline \hline
\multirow{5}{*}{X-Attn }  & 8 * (Self-Attn + X-Attn) & 276M & 5.6 \\ \cline{2-4} 
 & \textbf{2 * (Self-Attn + X-Attn)} & \textbf{175M} & \textbf{5.6} \\ \cline{2-4} 
& 1 * (Self-Attn + X-Attn) & 158M & 5.8 \\ \cline{2-4} 
 & 1 * X-Attn & 154M & 6.1 \\ \cline{2-4} 
 & 2 * RNN + 1 * X-Attn & 203M & 5.7 \\ \hline
%\begin{tabular}[c]{@{}l@{}}X-Attn + \\ unfreeze LM\end{tabular} & 2 * (Self-Attn + X-Attn) & 1.26B & 5.3 \\ \hline
\end{tabular}}
\end{table}

\subsection{Streaming latency-quality tradeoff}
\label{sec:lat-trade}

\begin{figure}[!tbh]
    \centering
    \includegraphics[width=0.8\linewidth]{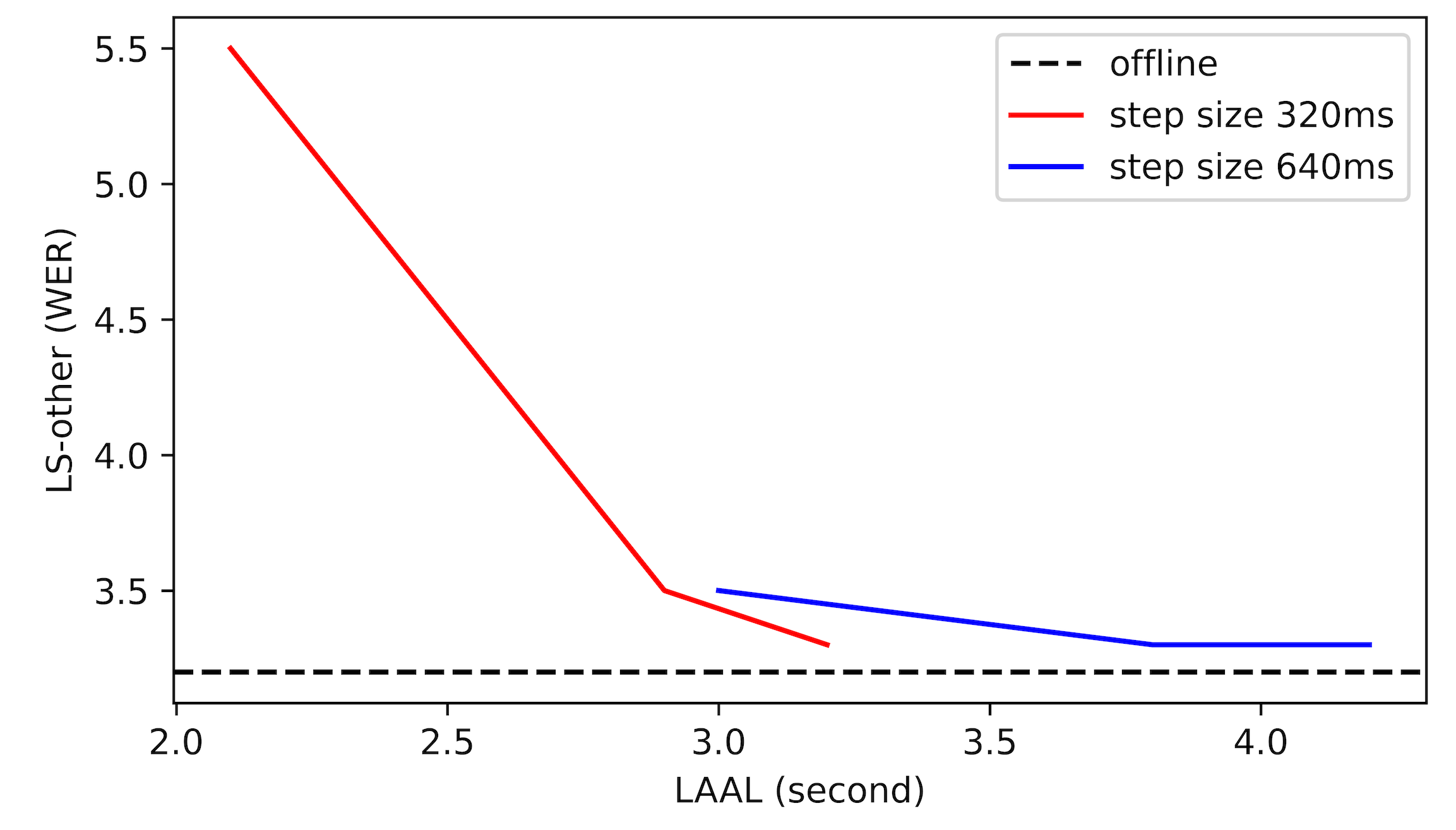}
        \vspace{-1.2em}
    \caption{Latency-Quality tradeoff curves of streaming ASR with BESTOW-S-bidi.}
    \label{fig:trade-asr}
    % /Users/zhehuaic/scatter_asr.py
    %https://docs.google.com/spreadsheets/u/2/d/1exc2yBn5l421OdGCHftzJLAqAFlEmjTzgt4lCohFPwE/edit?gid=139972364#gid=139972364
\end{figure}

\begin{figure}[!tbh]
    \centering
    \includegraphics[width=0.8\linewidth]{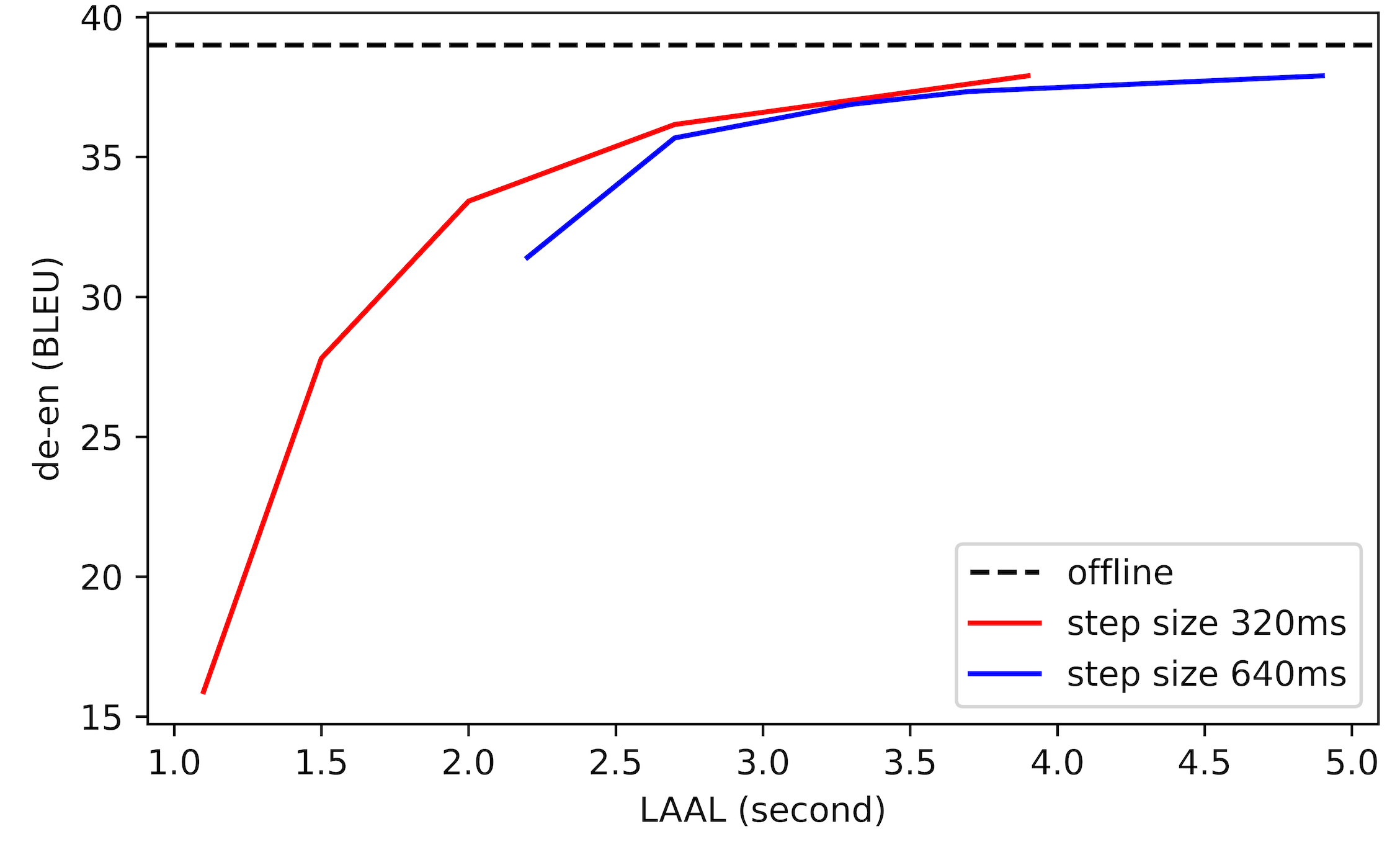}
     \vspace{-1.2em}
    \caption{Latency-Quality tradeoff curves of simultaneous speech translation with BESTOW-S-bidi.}
    \label{fig:trade-ast}
    % /Users/zhehuaic/scatter.py
    %https://docs.google.com/spreadsheets/u/2/d/1exc2yBn5l421OdGCHftzJLAqAFlEmjTzgt4lCohFPwE/edit?gid=139972364#gid=139972364
\end{figure}

To complete the streaming story of BESTOW-S, we give latency-quality tradeoff curves for streaming ASR and simultaneous
speech translation  in Figure~\ref{fig:trade-asr} and~\ref{fig:trade-ast} by varying $K$ in the inference time and reporting non-computation-aware laggings. The trend is in line with state-of-the-art bespoke systems in ASR~\cite{noroozi2023stateful} and AST~\cite{ma2023efficient} respectively. Optimizing computation-aware lagging will be interesting future topic.

%We further assess  the in-context learning ability by contextual word boosting through text prompting~\cite{chen2024salm} in Table~\ref{tab:ICL}. Comparing the original hypotheses and the word boosted hypotheses, the improvement is similar between the two architectures which shows in-context learning works on both.

%$\frac{(L_t+L_a)^2}{(L_t+L_a)(L_t)}=1+\frac{L_a}{L_t}$

\section{Conclusions}
\vspace{-0.5em}
In this work, 
we propose
 the first open SpeechLLM solution that enables {\em Streaming} and  {\em Multitask at scale} (beyond ASR) at the same time. %Moreover, the solution is end-to-end optimizable and allows LLM knowledge transfer to speech.
    The solution is based on a different backbone architecture from the popular
    Speech-LLaMA variants that is based on cross-attention and read-write policy. The novel backbone unifies the offline and streaming modes and achieves state-of-the-art on several large-scale and multitask  speech-to-text benchmarks, with {\em lower training/inference cost}.
We will release the code and checkpoints to promote next-generation SpeechLLM  using this backbone design. 
%A limitation of the model is that it might not work well on languages that are unseen during training, which will be our future work by leveraging in-context learning.

% \section{Acknowledgements}

\bibliographystyle{IEEEtran}
\bibliography{refs}

\begin{thebibliography}{10}
\providecommand{\url}[1]{#1}
\csname url@samestyle\endcsname
\providecommand{\newblock}{\relax}
\providecommand{\bibinfo}[2]{#2}
\providecommand{\BIBentrySTDinterwordspacing}{\spaceskip=0pt\relax}
\providecommand{\BIBentryALTinterwordstretchfactor}{4}
\providecommand{\BIBentryALTinterwordspacing}{\spaceskip=\fontdimen2\font plus
\BIBentryALTinterwordstretchfactor\fontdimen3\font minus \fontdimen4\font\relax}
\providecommand{\BIBforeignlanguage}[2]{{%
\expandafter\ifx\csname l@#1\endcsname\relax
\typeout{** WARNING: IEEEtran.bst: No hyphenation pattern has been}%
\typeout{** loaded for the language `#1'. Using the pattern for}%
\typeout{** the default language instead.}%
\else
\language=\csname l@#1\endcsname
\fi
#2}}
\providecommand{\BIBdecl}{\relax}
\BIBdecl

\bibitem{brown2020gpt}
T.~Brown \emph{et~al.}, ``Language models are few-shot learners,'' \emph{Advances in neural information processing systems}, vol.~33, pp. 1877--1901, 2020.

\bibitem{team2023gemini}
G.~Team, ``Gemini: a family of highly capable multimodal models,'' \emph{arXiv preprint arXiv:2312.11805}, 2023.

\bibitem{alayrac2022flamingo}
J.-B. Alayrac \emph{et~al.}, ``Flamingo: a visual language model for few-shot learning,'' \emph{Advances in Neural Information Processing Systems}, vol.~35, pp. 23\,716--23\,736, 2022.

\bibitem{liu2024llava}
H.~Liu, C.~Li, Q.~Wu, and Y.~J. Lee, ``Visual instruction tuning,'' \emph{Advances in neural information processing systems}, vol.~36, 2024.

\bibitem{zhang2023speechgpt}
D.~Zhang \emph{et~al.}, ``Speechgpt: Empowering large language models with intrinsic cross-modal conversational abilities,'' \emph{arXiv preprint arXiv:2305.11000}, 2023.

\bibitem{gong2023ltu-as}
Y.~Gong, A.~H. Liu, H.~Luo, L.~Karlinsky, and J.~Glass, ``Joint audio and speech understanding,'' in \emph{2023 IEEE Automatic Speech Recognition and Understanding Workshop (ASRU)}.\hskip 1em plus 0.5em minus 0.4em\relax IEEE, 2023, pp. 1--8.

\bibitem{tang2023salmonn}
C.~Tang \emph{et~al.}, ``Salmonn: Towards generic hearing abilities for large language models,'' \emph{arXiv preprint arXiv:2310.13289}, 2023.

\bibitem{wu2023decoder}
J.~Wu \emph{et~al.}, ``On decoder-only architecture for speech-to-text and large language model integration,'' in \emph{2023 IEEE Automatic Speech Recognition and Understanding Workshop (ASRU)}.\hskip 1em plus 0.5em minus 0.4em\relax IEEE, 2023, pp. 1--8.

\bibitem{fathullah2023prompting}
Y.~Fathullah, C.~Wu, E.~Lakomkin \emph{et~al.}, ``Prompting large language models with speech recognition abilities,'' \emph{arXiv:2307.11795}, 2023.

\bibitem{wang2023slm}
M.~Wang \emph{et~al.}, ``Slm: Bridge the thin gap between speech and text foundation models,'' in \emph{2023 IEEE Automatic Speech Recognition and Understanding Workshop (ASRU)}.\hskip 1em plus 0.5em minus 0.4em\relax IEEE, 2023, pp. 1--8.

\bibitem{qwen2023}
J.~Bai \emph{et~al.}, ``Qwen technical report,'' \emph{arXiv preprint arXiv:2309.16609}, 2023.

\bibitem{chen2024salm}
Z.~Chen \emph{et~al.}, ``Salm: Speech-augmented language model with in-context learning for speech recognition and translation,'' in \emph{ICASSP}.\hskip 1em plus 0.5em minus 0.4em\relax IEEE, 2024.

\bibitem{conneau2020unsupervised}
A.~Conneau \emph{et~al.}, ``Unsupervised cross-lingual representation learning for speech recognition,'' \emph{arXiv preprint arXiv:2006.13979}, 2020.

\bibitem{zhang2023google}
Y.~Zhang \emph{et~al.}, ``Google usm: Scaling automatic speech recognition beyond 100 languages,'' \emph{arXiv preprint arXiv:2303.01037}, 2023.

\bibitem{radford2022robust}
A.~Radford, J.~W. Kim, T.~Xu, G.~Brockman, C.~McLeavey, and I.~Sutskever, ``Robust speech recognition via large-scale weak supervision,'' 2022.

\bibitem{peng2024owsm}
Y.~Peng \emph{et~al.}, ``Owsm v3. 1: Better and faster open whisper-style speech models based on e-branchformer,'' \emph{arXiv preprint arXiv:2401.16658}, 2024.

\bibitem{barrault2023seamless}
L.~Barrault \emph{et~al.}, ``Seamless: Multilingual expressive and streaming speech translation,'' \emph{arXiv preprint arXiv:2312.05187}, 2023.

\bibitem{gulati2020conformer}
A.~Gulati, J.~Qin, C.-C. Chiu, N.~Parmar, Y.~Zhang, J.~Yu, W.~Han, S.~Wang, Z.~Zhang, Y.~Wu \emph{et~al.}, ``Conformer: Convolution-augmented transformer for speech recognition,'' in \emph{Interspeech}, 2020.

\bibitem{rekesh2023fast}
D.~Rekesh \emph{et~al.}, ``Fast conformer with linearly scalable attention for efficient speech recognition,'' in \emph{2023 IEEE Automatic Speech Recognition and Understanding Workshop (ASRU)}.\hskip 1em plus 0.5em minus 0.4em\relax IEEE, 2023, pp. 1--8.

\bibitem{conneau2022xtreme}
A.~Conneau \emph{et~al.}, ``Xtreme-s: Evaluating cross-lingual speech representations,'' \emph{arXiv preprint arXiv:2203.10752}, 2022.

\bibitem{shi2023ml}
J.~Shi, D.~Berrebbi, W.~Chen, H.-L. Chung, E.-P. Hu, W.~P. Huang, X.~Chang, S.-W. Li, A.~Mohamed, H.-y. Lee \emph{et~al.}, ``Ml-superb: Multilingual speech universal performance benchmark,'' \emph{arXiv preprint arXiv:2305.10615}, 2023.

\bibitem{rubenstein2023audiopalm}
P.~K. Rubenstein \emph{et~al.}, ``Audiopalm: A large language model that can speak and listen,'' \emph{arXiv preprint arXiv:2306.12925}, 2023.

\bibitem{kong2024audio}
Z.~Kong, A.~Goel, R.~Badlani, W.~Ping, R.~Valle, and B.~Catanzaro, ``Audio flamingo: A novel audio language model with few-shot learning and dialogue abilities,'' \emph{arXiv preprint arXiv:2402.01831}, 2024.

\bibitem{radhakrishnan2023whispering}
S.~Radhakrishnan \emph{et~al.}, ``Whispering llama: A cross-modal generative error correction framework for speech recognition,'' \emph{arXiv preprint arXiv:2310.06434}, 2023.

\bibitem{2020t5}
\BIBentryALTinterwordspacing
C.~Raffel \emph{et~al.}, ``Exploring the limits of transfer learning with a unified text-to-text transformer,'' \emph{Journal of Machine Learning Research}, vol.~21, no. 140, pp. 1--67, 2020. [Online]. Available: \url{http://jmlr.org/papers/v21/20-074.html}
\BIBentrySTDinterwordspacing

\bibitem{yu2023connecting}
W.~Yu \emph{et~al.}, ``Connecting speech encoder and large language model for asr,'' \emph{arXiv preprint arXiv:2309.13963}, 2023.

\bibitem{hussain2023m}
A.~S. Hussain, S.~Liu, C.~Sun, and Y.~Shan, ``$m^2$ugen: Multi-modal music understanding and generation with the power of large language models,'' \emph{arXiv preprint arXiv:2311.11255}, 2023.

\bibitem{zhang2020transformer}
Q.~Zhang \emph{et~al.}, ``Transformer transducer: A streamable speech recognition model with transformer encoders and rnn-t loss,'' in \emph{ICASSP}.\hskip 1em plus 0.5em minus 0.4em\relax IEEE, 2020.

\bibitem{noroozi2023stateful}
V.~Noroozi, S.~Majumdar, A.~Kumar, J.~Balam, and B.~Ginsburg, ``Stateful fastconformer with cache-based inference for streaming automatic speech recognition,'' \emph{arXiv preprint arXiv:2312.17279}, 2023.

\bibitem{graves2012sequence}
A.~Graves, ``Sequence transduction with recurrent neural networks,'' \emph{arXiv preprint arXiv:1211.3711}, 2012.

\bibitem{ma2020simulmt}
X.~Ma, J.~Pino, and P.~Koehn, ``Simulmt to simulst: Adapting simultaneous text translation to end-to-end simultaneous speech translation,'' \emph{arXiv preprint arXiv:2011.02048}, 2020.

\bibitem{ma2023efficient}
X.~Ma, A.~Sun, S.~Ouyang, H.~Inaguma, and P.~Tomasello, ``Efficient monotonic multihead attention,'' \emph{arXiv preprint arXiv:2312.04515}, 2023.

\bibitem{zhang2022information}
S.~Zhang and Y.~Feng, ``Information-transport-based policy for simultaneous translation,'' \emph{arXiv preprint arXiv:2210.12357}, 2022.

\bibitem{seide2024speech}
F.~Seide, M.~Doulaty, Y.~Shi, Y.~Gaur, J.~Jia, and C.~Wu, ``Speech reallm--real-time streaming speech recognition with multimodal llms by teaching the flow of time,'' \emph{arXiv preprint arXiv:2406.09569}, 2024.

\bibitem{tsunoo2024decoderonly}
E.~Tsunoo, H.~Futami, Y.~Kashiwagi, S.~Arora, and S.~Watanabe, ``Decoder-only architecture for streaming end-to-end speech recognition,'' 2024.

\bibitem{agostinelli2023simul}
V.~Agostinelli, M.~Wild, M.~Raffel, K.~A. Fuad, and L.~Chen, ``Simul-llm: A framework for exploring high-quality simultaneous translation with large language models,'' \emph{arXiv preprint arXiv:2312.04691}, 2023.

\bibitem{koshkin2024transllama}
R.~Koshkin, K.~Sudoh, and S.~Nakamura, ``Transllama: Llm-based simultaneous translation system,'' \emph{arXiv preprint arXiv:2402.04636}, 2024.

\bibitem{wang2023simultaneous}
M.~Wang, J.~Zhao, T.-T. Vu, F.~Shiri, E.~Shareghi, and G.~Haffari, ``Simultaneous machine translation with large language models,'' \emph{arXiv preprint arXiv:2309.06706}, 2023.

\bibitem{vaswani2017attention}
A.~Vaswani \emph{et~al.}, ``Attention is all you need,'' \emph{Advances in neural information processing systems}, vol.~30, 2017.

\bibitem{hu2021lora}
E.~J. Hu, Y.~Shen, P.~Wallis, Z.~Allen-Zhu, Y.~Li, S.~Wang, L.~Wang, and W.~Chen, ``Lora: Low-rank adaptation of large language models,'' \emph{arXiv preprint arXiv:2106.09685}, 2021.

\bibitem{panayotov2015librispeech}
V.~Panayotov, G.~Chen, D.~Povey, and S.~Khudanpur, ``Librispeech: an asr corpus based on public domain audio books,'' in \emph{ICASSP}.\hskip 1em plus 0.5em minus 0.4em\relax IEEE, 2015, pp. 5206--5210.

\bibitem{chen2021gigaspeech}
G.~Chen \emph{et~al.}, ``Gigaspeech: An evolving, multi-domain asr corpus with 10,000 hours of transcribed audio,'' \emph{arXiv preprint arXiv:2106.06909}, 2021.

\bibitem{wang2020covost}
C.~Wang, A.~Wu, and J.~Pino, ``Covost 2 and massively multilingual speech-to-text translation,'' \emph{arXiv preprint arXiv:2007.10310}, 2020.

\bibitem{data_gen_2024}
V.~Noroozi, Z.~Chen \emph{et~al.}, ``Instruction data generation and unsupervised adaptation for speech language models,'' in \emph{Interspeech}, 2024.

\bibitem{dynamicsuperb_leaderboard}
``Dynamic-superb leaderboard,'' \url{https://github.com/cyhuang-tw/dlhlp-dynamic-superb-leaderboard/blob/main/leaderboard.md}.

\bibitem{huang2023dynamic}
C.-y. Huang \emph{et~al.}, ``Dynamic-superb: Towards a dynamic, collaborative, and comprehensive instruction-tuning benchmark for speech,'' \emph{arXiv preprint arXiv:2309.09510}, 2023.

\bibitem{canary-1b}
``Canary-1b model,'' \url{https://huggingface.co/nvidia/canary-1b}, accessed: 2024-03-11.

\bibitem{bajaj2016msmacro}
P.~Bajaj \emph{et~al.}, ``Ms marco: A human generated machine reading comprehension dataset,'' \emph{arXiv preprint arXiv:1611.09268}, 2016.

\bibitem{papi2022over}
S.~Papi, M.~Gaido, M.~Negri, and M.~Turchi, ``Over-generation cannot be rewarded: Length-adaptive average lagging for simultaneous speech translation,'' \emph{arXiv preprint arXiv:2206.05807}, 2022.

\bibitem{kuchaiev2019nemo}
O.~Kuchaiev, J.~Li, H.~Nguyen \emph{et~al.}, ``{NeMo}: a toolkit for building ai applications using neural modules,'' \emph{arXiv:1909.09577}, 2019.

\bibitem{zhang2024tinyllama}
P.~Zhang, G.~Zeng, T.~Wang, and W.~Lu, ``Tinyllama: An open-source small language model,'' 2024.

\bibitem{librilight}
J.~{Kahn} \emph{et~al.}, ``Libri-light: A benchmark for asr with limited or no supervision,'' in \emph{ICASSP}, 2020, pp. 7669--7673, \url{https://github.com/facebookresearch/libri-light}.

\bibitem{commonvoice:2020}
R.~Ardila \emph{et~al.}, ``Common voice: A massively-multilingual speech corpus,'' in \emph{LREC}, 2020.

\bibitem{open-asr-leaderboard}
V.~Srivastav \emph{et~al.}, ``Open automatic speech recognition leaderboard,'' \url{https://huggingface.co/spaces/huggingface.co/spaces/open-asr-leaderboard/leaderboard}, 2023.

\bibitem{conneau2023fleurs}
A.~Conneau \emph{et~al.}, ``Fleurs: Few-shot learning evaluation of universal representations of speech,'' in \emph{SLT}.\hskip 1em plus 0.5em minus 0.4em\relax IEEE, 2023, pp. 798--805.

\bibitem{chan2016listen}
W.~Chan \emph{et~al.}, ``Listen, attend and spell: A neural network for large vocabulary conversational speech recognition,'' in \emph{ICASSP}.\hskip 1em plus 0.5em minus 0.4em\relax IEEE, 2016, pp. 4960--4964.

\end{thebibliography}

\end{document}